\documentclass[runningheads]{llncs}
\usepackage{graphicx}
\usepackage{xcolor}
\usepackage{fancyvrb}
\usepackage{hyperref}
\usepackage{xcolor}
\usepackage{footmisc}
\usepackage{lmodern}
\usepackage{ulem}

\hypersetup{colorlinks,allcolors=black}

\begin{document}
\title{KGTK: A Toolkit for Large Knowledge Graph Manipulation and Analysis}
\titlerunning{The Knowledge Graph Toolkit}
%
\author{Filip Ilievski\inst{1}\orcidID{0000-0002-1735-0686} \and
Daniel Garijo\inst{1}\orcidID{0000-0003-0454-7145} \and Hans Chalupsky\inst{1}\orcidID{0000-0002-8902-1662} \and Naren Teja Divvala\inst{1} \and Yixiang Yao\inst{1}\orcidID{0000-0002-2471-5591} \and Craig Rogers\inst{1}\orcidID{0000-0002-5818-3802} \and Rongpeng Li\inst{1}\orcidID{0000-0002-6911-8002} \and Jun Liu\inst{1} \and Amandeep Singh\inst{1}\orcidID{0000-0002-1926-6859} \and Daniel Schwabe\inst{2}\orcidID{0000-0003-4347-2940} \and Pedro Szekely\inst{1}}
\authorrunning{Filip Ilievski, Daniel Garijo et. al.}
%
\institute{Information Sciences Institute, University of Southern California \email{\{ilievski,dgarijo,hans,divvala,yixiangy,rogers, rli,junliu,amandeep,pszekely\}@isi.edu} \and Dept. of Informatics, Pontificia Universidade Cat\'olica Rio de Janeiro \\ \email{dschwabe@inf.puc-rio.br}
}

\maketitle              
\begin{abstract}

Knowledge graphs (KGs) have become the preferred technology for representing, sharing and adding knowledge to modern AI applications. 
While KGs have become a mainstream technology, the RDF/SPARQL-centric toolset for operating with them at scale is heterogeneous, difficult to integrate and only covers a subset of the operations that are commonly needed in data science applications. In this paper we present KGTK, a data science-centric toolkit designed to represent, create, transform, enhance and analyze KGs.
KGTK represents graphs in tables and leverages popular libraries developed for data science applications, enabling a wide audience of developers to easily construct knowledge graph pipelines for their applications.
We illustrate the framework with real-world scenarios where we have used KGTK to integrate and manipulate large KGs, such as Wikidata, DBpedia and ConceptNet.

\textbf{Resource type}: Software\\
\textbf{License}: MIT\\
\textbf{DOI}: https://doi.org/10.5281/zenodo.3828068\\
\textbf{Repository}: https://github.com/usc-isi-i2/kgtk/

\end{abstract}
%
%
%


\setcounter{footnote}{0}

\section{Introduction}

Knowledge graphs (KGs) have become the preferred technology for representing, sharing and using knowledge in applications. A typical use case is building a new knowledge graph for a domain or application by extracting subsets of several existing knowledge graphs, combining these subsets in application-specific ways, augmenting them with information from structured or unstructured sources, and computing analytics or inferred representations to support downstream applications.
For example, during the COVID-19 pandemic, several efforts focused on building KGs about scholarly articles related to the pandemic starting from the CORD-19 dataset provided by the Allen Institute for AI~\cite{Wang2020}.\footnote{\url{https://github.com/fhircat/CORD-19-on-FHIR/wiki/CORD-19-Semantic-Annotation-Projects}} Enhancing these data with with KGs such as DBpedia \cite{dbpedia} and Wikidata \cite{vrandecic_wikidata:_2014} to incorporate gene, chemical, disease and taxonomic information, and computing network analytics on the resulting graphs, requires the ability to operate these these KGs at scale.

Many tools exist to query, transform and analyze KGs. Notable examples include graph databases, such as 
RDF triple stores and Neo4J;\footnote{\url{https://neo4j.com}} tools for operating on RDF such as graphy\footnote{\url{https://graphy.link/}} and RDFlib\footnote{\url{https://rdflib.readthedocs.io/en/stable/}}, 
entity linking tools such as WAT~\cite{piccinno2014tagme} or BLINK~\cite{wu2019zero}, entity resolution tools such as MinHash-LSH~\cite{leskovec2020mining} or MFIBlocks~\cite{kenig2013mfiblocks}, libraries to compute graph embeddings such as PyTorch-BigGraph~\cite{lerer2019pytorch} and libraries for graph analytics, such as graph-tool\footnote{\url{https://graph-tool.skewed.de/}} and NetworkX.\footnote{\url{https://networkx.github.io/}} 

There are three main challenges when using these tools together. First, tools may be challenging to set up with large KGs (e.g., the Wikidata RDF dump takes over a week to load into a triple store) and often need custom configurations that require significant expertise. Second, interoperating between tools requires developing data transformation scripts, as some of them may not support the same input/output representation. Third, composing two or more tools together (e.g., to filter, search, and analyze a KG) includes writing the intermediate results to disk, which is time and memory consuming for large KGs. 


In this paper, we introduce the Knowledge Graph Toolkit (KGTK), a framework for manipulating, validating, and analyzing large-scale KGs. Our work is inspired by Scikit-learn~\cite{scikit-learn} and SpaCy,\footnote{\url{https://spacy.io/}} two popular toolkits for machine learning and natural language processing that have had a vast impact by making these technologies accessible to data scientists and software developers. KGTK aims to build a comprehensive library of tools and methods to enable easy composition of KG operations (validation, filtering, merging, centrality, text embeddings, etc.) to build knowledge-based AI applications. The contributions of KGTK are:

\begin{itemize}
\item The \textbf{KGTK file format}, which allows representing KGs as hypergraphs. This format unifies the Wikidata data model \cite{vrandecic_wikidata:_2014} based on items, claims, qualifiers, and references; property graphs that support arbitrary attributes on nodes and edges; RDF-Schema-based graphs such as DBpedia \cite{dbpedia}; and general purpose RDF graphs with various forms of reification. The KGTK format uses tab-separated values (TSV) to represent edge lists, making it easy to process with many off-the-shelf tools.

\item A comprehensive \textbf{validator and data cleaning} module to verify compliance with the KGTK format, and normalize literals like strings and numbers.
\item \textbf{Import modules} to transform different formats into KGTK, including N-Triples \cite{Seaborne:14:RN}, Wikidata qualified terms, and ConceptNet \cite{speer2016conceptnet}.
\item \textbf{Graph manipulation modules} for bulk operations on graphs to validate, clean, filter, join, sort, and merge KGs. Several of these are implemented as wrappers of common, streaming Unix tools like awk\footnote{\url{https://linux.die.net/man/1/awk}}, sort, and join. 
\item \textbf{Graph querying and analytics modules} to compute centrality measures, connected components, and text-based graph \textbf{embeddings} using state-of-the-art language models: RoBERTa~\cite{liu2019roberta}, BERT~\cite{devlin2018bert}, and DistilBERT~\cite{sanh2019distilbert}. Common queries, such as computing the set of nodes reachable from other nodes, are also supported.
\item \textbf{Export modules} to transform KGTK format into diverse standard and commonly used formats, such as RDF (N-Triples), property graphs in Neo4J format, and GML to invoke tools such as graph-tool or Gephi.\footnote{\url{https://gephi.org/}}  
\item A \textbf{framework for composing multiple KG operations}, based on Unix pipes. The framework uses the KGTK file format on the standard input and output to combine tools written in different programming languages.
\end{itemize}

KGTK provides an implementation that integrates all these methods relying on widely used tools and standards, thus allowing their composition in pipelines to operate with large KGs like Wikidata on an average laptop. 

The rest of the paper is structured as follows. Section \ref{sec:scenario} describes a motivating scenario and lists the requirements for a graph manipulation toolkit. Section \ref{sec:kgtk} describes KGTK by providing an overview of its file format, supported operations, and examples on how to compose them together. Next, Section \ref{sec:showcase} showcases how we have used KGTK on three different real-world use cases, together with the current limitations of our approach. We then review relevant related work in Section \ref{sec:relWork}, and we conclude the paper in Section \ref{sec:conclusions}.

\section{Motivating Scenario}
\label{sec:scenario}



The 2020 coronavirus pandemic led to a series of community efforts to publish and share common knowledge about COVID-19 using KGs. Many of these efforts use the COVID-19 Open Research Dataset (CORD-19) \cite{Wang2020}, compiled by the Allen Institute for AI. CORD-19 is a free resource containing over 44,000 scholarly articles, including over 29,000 with full text, about COVID-19 and the coronavirus family of viruses. Having an integrated KG would allow easy access to information published in scientific papers, as well as to general medical knowledge on genes, proteins, drugs, and diseases mentioned in these papers, and their interactions.

In our work, we integrated the CORD-19 corpus with gene, chemical, disease, and taxonomic knowledge from Wikidata and CTD databases,\footnote{\url{http://ctdbase.org/}} as well as entity extractions from Professor Heng Ji's BLENDER lab at UIUC.\footnote{\url{https://blender.cs.illinois.edu/}} We extracted all the items and statements for the 30,000 articles in the CORD-19 corpus \cite{Wang2020} that were present in Wikidata at the time of extraction, added all Wikidata articles, authors, and entities mentioned in the BLENDER corpus, homogenized the data to fix inconsistencies (e.g., empty values), created nodes and statements for entities that were absent in Wikidata, incorporated metrics such as PageRank for each KG node, and exported the output in both RDF and Neo4J. 

This use case exhibited several of the challenges that KGTK is designed to address. For example, extracting a subgraph from Wikidata articles is not feasible using SPARQL queries as it would have required over 100,000 SPARQL queries; using RDF tools on the Wikidata RDF dump (107 GB compressed) is difficult because its RDF model uses small graphs to represent each Wikidata statement; using the Wikidata JSON dump is possible, but requires writing custom code as the schema is specific to Wikidata (hence not reusable for other KGs). In addition, while graph-tool allowed us to compute graph centrality metrics, its input format is incompatible with RDF, requiring a transformation. 


Other efforts employed a similar set of processing steps~\cite{Wang2020}.\footnote{A list of such projects can be found in \url{https://github.com/fhircat/CORD-19-on-FHIR/wiki/CORD-19-Semantic-Annotation-Projects}} These range from mapping the CORD-19 data to RDF,\footnote{\url{https://github.com/nasa-jpl-cord-19/covid19-knowledge-graph}, \url{https://github.com/GillesVandewiele/COVID-KG/}} to adding annotations to the articles in the dataset pointing to entities extracted from the text, obtained from various sources \cite{gmg20}.\footnote{\url{http://pubannotation.org/collections/CORD-19}}
A common thread among these efforts involves leveraging existing KGs such as Wikidata and Microsoft Academic Graph
to, for example, build a citation network of the papers, authors, affiliations, etc.\footnote{\url{https://scisight.apps.allenai.org/clusters}\label{scisight}} Other efforts focused on extraction of relevant entities (genes, proteins, cells, chemicals, diseases), relations (causes, upregulates, treats, binds), and 
linking them to KGs such as Wikidata and DBpedia.
Graph analytics operations followed, such as computing centrality measures in order to support identification of key articles, people or substances,\footref{scisight} or generation of various embeddings to recommend relevant literature associated with an entity.\footnote{\url{https://github.com/vespa-engine/cord-19/blob/master/README.md}}
The resulting graphs were deployed as SPARQL endpoints, or exported as RDF dumps, CSV, or JSON files.


These examples illustrate the need for composing sequences of integrated KG operations that extract, modify, augment and analyze knowledge from existing KGs, combining it with non-KG datasets to produce new KGs. Existing KG tools do not allow users to seamlessly run such sequences of graph manipulation tasks in a pipeline. We propose that an effective toolkit that supports the construction of modular KG pipelines has to meet the following criteria:

\begin{enumerate}
\item \textbf{A simple representation format} that all modules in the toolkit operate on (the equivalent of \textit{datasets} in Scikit-learn and \textit{document model} in SpaCy), to enable tool integration without additional data transformations.  

\item \textbf{Ability to incorporate mature existing tools}, wrapping them to support a common API and input/output format. The scientific community has worked for many years on efficient techniques for manipulation of graph and structured data. The toolkit should be able to accommodate them without the need for a new implementation. 
 
\item \textbf{A comprehensive set of features} that include import and export modules for a wide variety of KG formats, modules to select, transform, combine, link, and merge KGs, modules to improve the quality of KGs and infer new knowledge, and modules to compute embeddings and graph statistics. Such a rich palette of functionalities would largely support use cases such as the ones presented in this section.

\item A \textbf{pipeline mechanism} to allow composing modules in arbitrary ways to process large public KGs such as Wikidata, DBpedia, or ConceptNet.
\end{enumerate}



\section{KGTK: The Knowledge Graph Toolkit}\label{sec:kgtk} 

\begin{figure}[t!]
    \centering
    \includegraphics[width=0.97\textwidth]{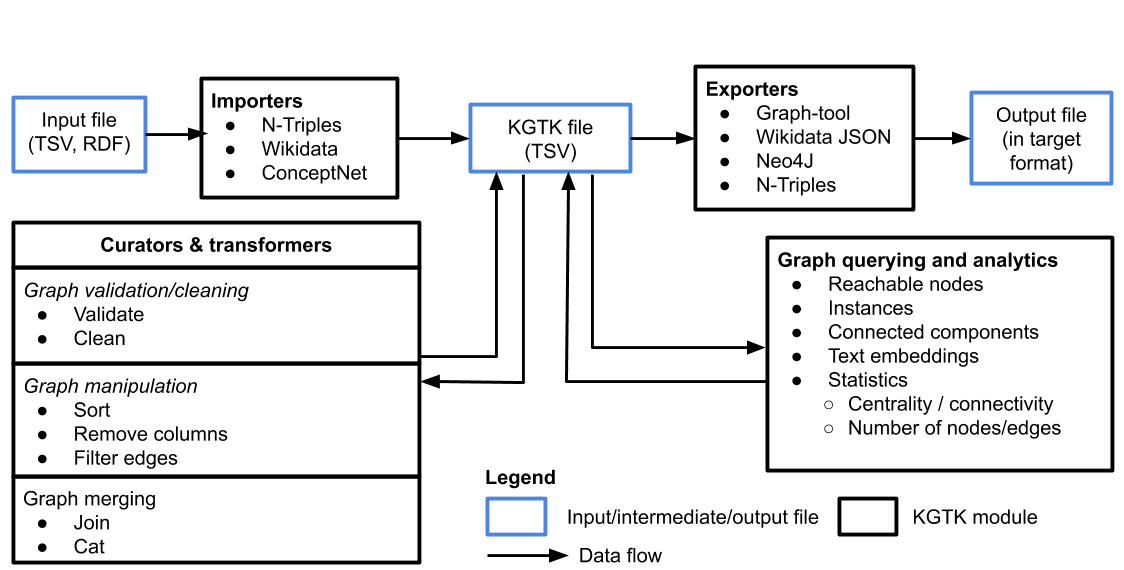}
    \caption{Overview of the usage workflow and features included in KGTK.}
    \label{fig:overview}
\end{figure}

KGTK helps manipulating, curating, and analyzing large real-world KGs, in which each statement may have multiple qualifiers such as the statement source, its creation date or its measurement units. Figure \ref{fig:overview} shows an overview of the different capabilities of KGTK. Given an input file with triples (either as tab-separated values, Wikidata JSON, or N-Triples), we convert it to an internal representation (the \textit{KGTK file format}, described in Section \ref{sec:format}) that we use as main input/output format for the rest of the features included in the toolkit. Once data is in KGTK format, we can perform operations for curating (data validation and cleaning), transforming (sort, filter, or join) and analyzing the contents of a KG (computing embeddings, statistics, node centrality). KGTK also provides export operations to common formats, such as N-Triples, Neo4J, and JSON. 
The KGTK operations are described in Section \ref{sec:features}, whereas their composition into pipelines is illustrated in Section \ref{sec:chain}.

\subsection{KGTK file format}\label{sec:format}
\label{ssec:kgtk-format}


KGTK uses a tab-separated column-based text format to describe any attributed, labeled or unlabeled hypergraph. We chose this format instead of an RDF serialization for three reasons. First, tabular formats are easy to generate and parse by standard tools. Second, this format is self-describing and easy to read by humans. Finally, it provides a simple mechanism to define hypergraphs and edge qualifiers, which may be more complicated to describe using Turtle or JSON. 


KGTK defines KGs as a set of nodes and a set of edges between those nodes. All concepts of meaning are represented via an edge, including edges themselves, allowing KGTK to represent generalized hypergraphs (while supporting the representation of RDF graphs). 
The snippet below shows a simple example of a  KG in KGTK format with three people (\textit{Moe}, \textit{Larry} and \textit{Curly}), the creator of the statements (\textit{Hans}) and the original source of the statements (\textit{Wikipedia}):  
{\footnotesize
\vbox{
\begin{Verbatim}[commandchars=\\\{\}]
\textbf{
    node1     label       node2    creator   source      id}
    "Moe"     rdf:type    Person   "Hans"    Wikipedia   E1
    "Larry"   rdf:type    Person   "Hans"    Wikipedia   E2
    "Curly"   rdf:type    Person             Wikipedia   E3
    "Curly"   hasFriend   "Moe"              Wikipedia   E4
\end{Verbatim}
}
}
The first line of a KGTK file declares the headers for the document. The reserved words \textit{node1}, \textit{label} and \textit{node2} are used to describe the subject, property and object being described, while \textit{creator} and \textit{source} are optional qualifiers for each statement that provide additional provenance information about the creator of a statement and the original source. Note that the example is not using namespace URIs for any nodes and properties, as they are not needed for local KG manipulation. Still, namespace prefixes (e.g., \texttt{rdf}) may be used for mapping back to RDF after the KG manipulations with KGTK. Nodes and edges have unique IDs (when IDs are not present, KGTK generates them automatically).

The snippet below illustrates the representation of qualifiers for individual edges, and shows how the additional columns in the previous example may be represented as edges about edges:

{\footnotesize
\vbox{
\begin{Verbatim}[commandchars=\\\{\}]
\textbf{
    node1     label       node2       id}
    "Moe"     rdf:type    Person      E1
    E1        source      Wikipedia   E5
    E1        creator     "Hans"      E6
    "Larry"   rdf:type    Person      E2  
\end{Verbatim}
}
}
KGTK is designed to support commonly-used typed literals:
\begin{itemize}

    \item \textbf{Language tags}: represented following a subset of the RDF convention, language tags are two- or three-letter ISO 639-3 codes, optionally followed by a dialect or location subtag. 
    Example: \verb+`Sprechen sie deutsch?'@de+.  
    
    \item \textbf{Quantities}: represented using a variant of the Wikidata format \texttt{amount}$\sim$ \texttt{toleranceUxxxx}.  A quantity starts with an \textit{amount} (number), followed by an optional \textit{tolerance interval}, followed by either a combination of \textit{standard (SI) units} or a \textit{Wikidata node} defining the unit (e.g., Q11573 indicates ``meter''). Examples include \verb+10m+, \verb$-1.2e+2[-1.0,+1.0]kg.m/s2$ or \verb$+17.2Q494083$

    \item \textbf{Coordinates}: represented by using the Wikidata format @LAT/LON, for example: \verb+@043.26193/010.92708+
    
    \item \textbf{Time literals}: represented with a $^\wedge$ character (indicating the tip of a clock hand) and followed by an ISO 8601 date and an optional precision designator, for example: \verb$^1839-00-00T00:00:00Z/9$
\end{itemize}


The full KGTK file format specification is available online.\footnote{\url{https://kgtk.readthedocs.io/en/latest/specification/}} 




\subsection{KGTK Operations}\label{sec:features}

KGTK currently supports 13 operations (depicted in Figure \ref{fig:overview}),\footnote{\url{https://kgtk.readthedocs.io/en/latest}} grouped into four modules: importing modules, graph manipulation modules, graph analytics modules, and exporting modules. We describe each of these modules below.

\subsubsection{Importing and exporting from KGTK}~\\



\textbf{1.}
The \texttt{import} operation transforms an external graph format into KGTK TSV format.
KGTK supports importing a number of data formats, including N-Triples, ConceptNet, and Wikidata (together with qualifiers). 

\textbf{2. } The \texttt{export} operation transforms a KGTK-formatted graph to a wide palette of formats: TSV (by default), N-Triples, Neo4J Property Graphs, graph-tool and the Wikidata JSON format.

\subsubsection{Graph curation and transformation}~\\


\textbf{3. } 
The \texttt{validate} operation ensures that the data meets the KGTK file format specification, detecting errors such as nodes with empty values, values of unexpected length (either too long or too short), potential errors in strings (quotation errors, incorrect use of language tags, etc.), incorrect values in dates, etc. Users may customize the parsing of the file header, each line, and the data values, as well as choose the action taken when a validation rule fails.

\textbf{4.} The \texttt{clean} operation fixes a substantial number of errors detected by \texttt{validate}, by correcting some common mistakes in data encoding (such as not escaping `pipe' characters), 
replacing invalid dates, normalizing values for quantities, languages and coordinates using the KGTK convention for literals. Finally, it removes rows that still do not meet the KGTK specification (e.g., rows with empty values for required columns or rows with an invalid number of columns).


\textbf{5.}  \texttt{sort} efficiently reorders any KGTK file according to one or multiple columns. 
\texttt{sort} is useful to organize edge files so that, for example, all edges for \texttt{node1} are contiguous, enabling efficient processing in streaming operations.


\textbf{6.} The \texttt{remove-columns} operation 
removes a subset of the columns in a KGTK file (\texttt{node1} (source), \texttt{node2} (object), and \texttt{label} (property) cannot be removed). 
This is useful in cases where columns have lengthy values and are not relevant to the use case pursued by a user, e.g., removing edge and graph identifiers when users aim to compute node centrality or calculate embeddings.



\textbf{7.} The \texttt{filter} operation selects edges from a KGTK file, by specifying constraints (``patterns'') on the values for node1, label, and node2. The \texttt{pattern} language, inspired by \url{graphy.js}, has the following form: ``\texttt{subject-pattern ; predicate-pattern ; object-pattern}".
For each of the three columns, the filtering pattern can consist of a list of symbols separated using commas. Empty patterns indicate that no filter should be performed for a column. For instance, to select all edges that have property P154 or P279, we can use the pattern `` ; P154,P279 ; ''. Alternatively, a common query of retrieving edges for all humans from Wikidata corresponds to the filter `` ; P31 ; Q5''.

\textbf{8.} The \texttt{join} operation will join two KGTK files. 
Inner join, left outer join, right outer join, and full outer join are all supported. 
When a join takes place, the columns from two files are merged into the set of columns for the output file.
By default, KGTK will join based on the \texttt{node1} column, although it can be configured to join by edge \texttt{id}. 
KGTK also allows the \texttt{label} and \texttt{node2} columns to be added to the join. 
Alternatively, the user may supply a list of join columns for each file giving them full control over the semantics of the result.


\textbf{9.} The \texttt{cat} operation concatenates any number of files into a single, KGTK-compliant graph file. 




\subsubsection{Graph querying and analytics}~\\


\textbf{10.} \texttt{reachable-nodes}: given a set of nodes N and a set of properties P, this operation computes the set of reachable nodes $R$ that contains the nodes that can be reached from a node $n \in N$ via paths containing any of the properties in P. This operation can be seen as a (joint) closure computation over one or multiple properties for a predefined set of nodes. A common application of this operation is to compute a closure over the subClassOf property, which benefits downstream tasks such as entity linking or table understanding. 



\textbf{11.} The \texttt{connected-components} operation finds all connected components (communities) in a graph (e.g., return all the communities connected via an \texttt{owl:sameAs} edge in a KGTK file).



\textbf{12. } The \texttt{text-embeddings} operation computes embeddings for all nodes in a graph by computing a sentence embedding over a lexicalization of the neighborhood of each node. The lexicalized sentence is created based on a template whose simplified version is:
{\footnotesize
\begin{verbatim}
{label-properties}, {description-properties} is a {isa-properties},
has {has-properties}, and {properties:values}.
\end{verbatim}
}

\sloppy
The properties for \texttt{label-properties}, \texttt{description-properties}, \texttt{isa-properties}, \texttt{has-properties}, and \texttt{property-values} pairs are specified as input arguments to the operation. An example sentence is ``Saint David, patron saint of Wales is a human, Catholic priest, Catholic bishop, and has date of death, religion, canonization status, and has place of birth Pembrokeshire''. The sentence for each node is encoded into an embedding using one of 16 currently supported variants of three state-of-the-art language models: BERT, DistilBERT, and RoBERTa. Computing similarity between such entity embeddings is a standard component of modern decision making systems such as entity linking, question answering, or table understanding.

\textbf{13.} The \texttt{graph-statistics} operation computes various graph statistics and centrality metrics. The operation generates a graph summary, containing its number of nodes, edges, and most common relations. In addition, it can compute graph degrees, HITS centrality, and PageRank values. Aggregated statistics (minimum, maximum, average, and top nodes) for these connectivity/centrality metrics are included in the summary, whereas the individual values for each node are represented as edges in the resulting graph. The graph is assumed to be directed, unless indicated differently.






\subsection{Composing operations into pipelines}\label{sec:chain}

KGTK has a pipelining architecture based on Unix pipes\footnote{\url{https://linux.die.net/man/7/pipe}} that allows chaining most operations introduced in the previous section by using the standard input/output and the KGTK file format. 
Pipelining increases efficiency by avoiding the need to write files to disk and supporting parallelism allowing downstream commands to process data before upstream commands complete.
We illustrate the chaining operations in KGTK with three examples from our own work. Note that we have implemented a shortcut pipe operator ``/'', which allows users to avoid repeating \texttt{kgtk} in each of their operations. For readability, command arguments are slightly simplified in the paper. Jupyter Notebooks that implement these and other examples can be found online.\footnote{\url{https://github.com/usc-isi-i2/kgtk/tree/master/examples}}   


\textbf{Example 1}: Alice wants to import the English subset of ConceptNet \cite{speer2016conceptnet} in KGTK format to extract a filtered subset where two concepts are connected with a more precise semantic relation such as \texttt{/r/Causes} or \texttt{/r/UsedFor} (as opposed to weaker relations such as \texttt{/r/RelatedTo}). For all nodes in this subset, she wants to compute text embeddings and store them in a file called \texttt{emb.txt}.

To extract the desired subset, the sequence of KGTK commands is as follows:
{\footnotesize
\begin{verbatim}
kgtk import-conceptnet --english_only conceptnet.csv / \
  filter -p "; /r/Causes,/r/UsedFor,/r/Synonym,/r/DefinedAs,/r/IsA ;" / \
  sort -c 1,2,3 > sorted.tsv
\end{verbatim}
}

To compute embeddings for this subset, she would use \texttt{text-embeddings}:
{\footnotesize
\begin{verbatim}
kgtk text-embeddings --label-properties "/r/Synonym" \
  --isa-properties "/r/IsA" --description-properties "/r/DefinedAs" \
  --property-value "/r/Causes" "/r/UsedFor" \ 
  --model bert-large-nli-cls-token -i sorted.tsv \
  > emb.txt
\end{verbatim}
}

\textbf{Example 2}: Bob wants to extract a subset of Wikidata that contains only edges of the `member of' (P463) property, and strip a set of columns that are not relevant for his use case (\texttt{\$ignore\_col}), such as id and rank. While doing so, Bob would also like to clean any erroneous edges. On the clean subset, he would compute graph statistics, including PageRank values and node degrees. Here is how to perform this functionality in KGTK (after Wikidata is already converted to a KGTK file called \texttt{wikidata.tsv} by \texttt{import-wikidata}):
{\footnotesize
\begin{verbatim}
kgtk filter -p ' ; P463 ; ' /  clean_data / 
    remove-columns -c "$ignore_cols" wikidata.tsv > graph.tsv
kgtk graph-statistics --directed --degrees --pagerank graph.tsv
\end{verbatim}
}

\textbf{Example 3}: Carol would like to concatenate two subsets of Wikidata: one containing occupations for several notable people: Sting, Roger Federer, and Nelson Mandela; and the other containing all `subclass of' (P279) relations in Wikidata. The concatenated file needs to be sorted by subject, after which she would compute the set of reachable nodes for these people via the properties `occupation' (P106) or `subclass of' (P279). To achieve this in KGTK, Carol first needs to extract the two subsets with the \texttt{filter} operation:
{\footnotesize
\begin{verbatim}
kgtk filter -p 'Q8023,Q483203,Q1426;P106;' wikidata.tsv > occupation.tsv
kgtk filter -p ` ; P279 ; ' wikidata.tsv > subclass.tsv
\end{verbatim}
}

Then, she can merge the two files into one, sort it, and compute reachability:
{\footnotesize
\begin{verbatim}
kgtk cat occupation.tsv subclass.tsv / \
     sort -c node1 > sorted.tsv
kgtk reachable-nodes --props P106,P279 --root "Q8023,Q483203,Q1426" \
     sorted.tsv > reachable.tsv
\end{verbatim}
}



\section{Discussion}\label{sec:showcase}




Validating, merging, transforming and analyzing KGs at scale is an open challenge for knowledge engineers, and even more so for data scientists. Complex SPARQL queries often time out on online endpoints, while working with RDF dumps locally takes time and expertise. In addition, popular graph analysis tools do not operate with RDF, making analysis complex for data scientists.



The KGTK format intentionally does not distinguish attributes or qualifiers of nodes and edges from full-fledged edges. Tools operating on KGTK graphs can instead interpret edges differently when desired.  In the KGTK file format, everything can be a node, and every node can have any type of edge to any other node. To do so in RDF requires adopting more complex mechanisms such as reification, typically leading to efficiency issues. This generality allows KGTK files to be mapped to most existing DBMSs, and to be used in powerful data transformation and analysis tools such as Pandas.\footnote{\url{https://pandas.pydata.org}} 

We believe KGTK will have a significant impact
within and beyond the Semantic Web community by helping users to easily perform usual data science operations on large KGs. To give an idea, we downloaded Wikidata (truthy statements distribution, 23.2GB\footnote{\url{https://dumps.wikimedia.org/wikidatawiki/entities/latest-truthy.nt.bz2}}) and performed a test of filtering out all Qnodes (entities) which have the P31 property (instance of) in Wikidata. This filter took over 20 hours in Apache Jena and RDFlib. In graphy, the time went down to 4h 15min. Performing the same operation in KGTK took less than 1h 30min. 

\begin{figure}[t!]
    \centering
    \includegraphics[width=\textwidth]{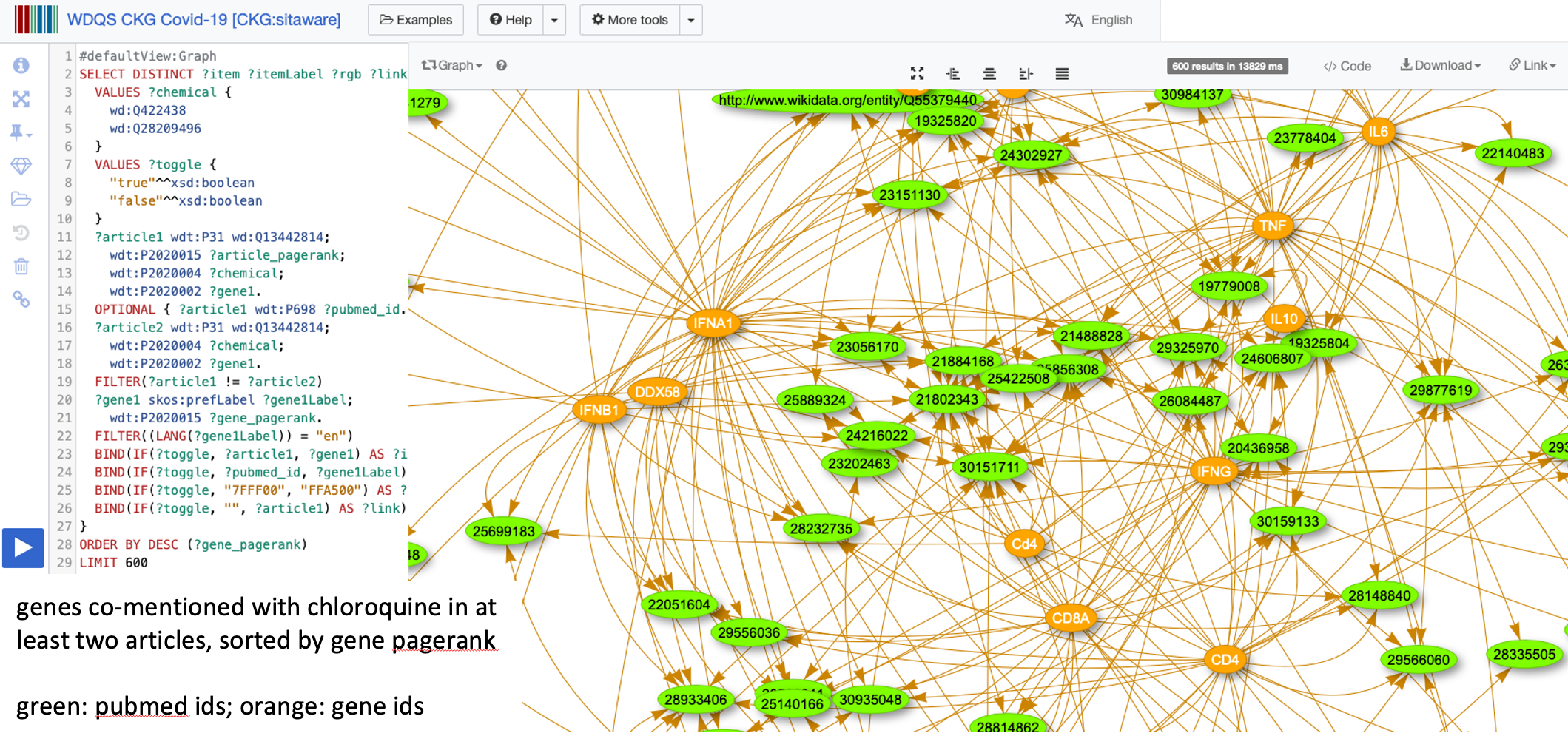}
    \caption{SPARQL query and visualization of the CORD-19 use case, illustrating the use of the Wikidata infrastructure using our KG that includes a subset of Wikidata augmented with new properties such as ``mentions gene'' and ``pagerank''.}
    \label{fig:covid-visualization}
\end{figure}
We have been using KGTK in our own work to integrate and analyze KGs:
\begin{itemize}
    \item \textbf{CORD-19}: As described in Section \ref{sec:scenario}, we used KGTK to combine extracted information from the papers in the CORD-19 dataset (such as entities of interest) with metadata about them, and general medical and biology knowledge, all found in Wikidata, CTD and the BLENDER datasets. A notebook illustrating the operations used in this use case is available online.\footnote{\url{https://github.com/usc-isi-i2/CKG-COVID-19/blob/dev/build-covid-kg.ipynb}} 
    Figure~\ref{fig:covid-visualization} shows the CORD-19 KGTK KG loaded in Wikidata SPARQL query interface. The KGTK tools exported the CORD-19 KG to RDF triples in a format compatible with Wikidata. 

\item \textbf{Commonsense Knowledge Graph (CSKG)}: Commonsense knowledge is dispersed across a number of (structured) knowledge sources, like ConceptNet and ATOMIC~\cite{sap2019atomic}. After consolidating several such sources into a single CSKG~\cite{ilievski2020consolidating}, we used KGTK to compute graph statistics (e.g., number of edges or most frequent relations), HITS, PageRank, and node degrees, in order to measure the impact of the consolidation on the graph connectivity and centrality. We also created RoBERTa-based embeddings of the CSKG nodes, which we are currently using for downstream question answering. A notebook illustrating the operations in this use case is available online.\footnote{\url{https://github.com/usc-isi-i2/kgtk/blob/master/examples/CSKG.ipynb}} 


\item \textbf{Integrating and exporting Ethiopian quantity data:}
We are using KGTK to create a custom extension of Wikidata with data about Ethiopia,\footnote{\url{https://datamart-upload.readthedocs.io/en/latest/REST-API-tutorial/}} by integrating quantity indicators like crime, GDP, population, etc.
\end{itemize}





The heterogeneity of these cases shows how KGTK can be adopted for multi-purpose data-science operations over KGs, independently of the domain. The challenges described in them are common in data integration and data science. Given the rate at which KGs are gaining popularity, we expect KGTK to fill a key gap faced by many practitioners wanting to use KGs in their applications.

The primary limitation of KGTK lies in its functionality coverage. The main focus so far has been on supporting basic operations for manipulating KGs, and therefore KGTK does not yet incorporate powerful browsing and visualization tools, or advanced tools for KG identification tasks such as link prediction, entity resolution, and ontology mapping.
Since KGTK is proposed as a new resource, we have no usage metrics at the time of writing this paper.

\section{Related Work}\label{sec:relWork}
Many of the functionalities in KGTK for manipulating and transforming KGs (i.e., join operations, filtering entities, general statistics, and node reachability) can be translated into queries in SPARQL. However, the cost of these queries over large endpoints is often too high, and they will time out or take too long to produce a response. In fact, many SPARQL endpoints have been known to have limited availability and slow response times for many queries~\cite{buil2013sparql}, leaving no choice but to download their data locally for any major KG manipulation. 
Additionally, it is unclear how to extend SPARQL to support functionalities  such as computing embeddings or node centrality.
A scalable alternative to SPARQL is Linked Data Fragments (LDF)~\cite{verborgh2014web}. The list of native operations in LDF boils down to triple pattern matching, resembling our proposed \texttt{filter} operation. However, operations like merging and joining are not trivial in LDF, while more complex analytics and querying, like embedding computation, are not supported. 

Other works have proposed offline querying. LOD Lab~\cite{beek2016lod} and LOD-a-lot~\cite{fernandez2017lod}  combine LDF with an efficient RDF compression format, called Header Dictionary Triples (HDT)~\cite{martinez2012exchange,fernandez2018hdtq}, in order to store a LOD dump of 30-40B statements. Although the LOD Lab project also employed mature tooling, such as Elastic Search and bash operations, to provide querying over the data, the set of available operations is restricted by employing LDF as a server, as native LDF only supports pattern matching queries. 
The HDT compression format has also been employed by other efforts, such as sameAs.cc \cite{beek2018sameas}, which performs closure and clustering operations over half a billion identity (same-as) statements. However, HDT cannot be easily used by existing tools (e.g., graph-tool or pandas), and it does not describe mechanisms for supporting qualifiers (except for using reification on statements, which complicates the data model). 

The recent developments towards supporting triple annotations with  RDF*~\cite{hartig2017rdf} provide support for qualifiers; yet, this format is still in its infancy and we expect it to inherit the challenges of RDF, as described before.


Several RDF libraries exist for different programming languages, such as RDFLib in Python, graphy in JavaScript, and Jena in Java. The scope of these libraries is different from KGTK, as they focus on providing the building blocks for creating RDF triples, rather than a set of operators to manipulate and analyze large KGs (validate, merge, sort, statistics, etc.).



Outside of the Semantic Web community, prominent efforts perform graph operations in graph databases like Neo4J or libraries like graph-tool, which partially overlap with the operations included in KGTK. We acknowledge the usefulness of these tools for tasks like pattern matching and graph traversal, and therefore we provide an export to their formats to enable users to take advantage of those capabilities. However, these tools also have limitations. First, Neo4J ``only allows one value per attribute property'' and it ``does not currently support queries involving joins or lookups on any information associated with edges, including edge ids, edge types, or edge attributes''~\cite{hernandez2016querying}. The KGTK representation does not have these limitations, and the tasks above can be performed using KGTK commands or via export to SPARQL and graph-tool. Second, while Neo4J performs very well on traversal queries, it is not optimized to run on bulk, relational queries, like ``who are the latest reported sports players?'' Similarly,~\cite{hernandez2016querying} shows that Neo4J performs worse and times out more frequently than Postgres and Virtuoso on atomic queries and basic graph patterns, even after removing the labels to improve efficiency. KGTK supports bulk and simple table queries, complex queries are handled by exporting to RDF and Postgres. 

Graph-tool provides rich functionality and can natively support property graphs. However, it needs to be integrated with other tools for operations like computation of embeddings or relational data operations, requiring additional expertise. 

Finally, the KGX toolkit\footnote{\url{https://github.com/NCATS-Tangerine/kgx}} has a similar objective as KGTK, but it is scoped to process KGs aligned with the Biolink Model, a datamodel describing biological entities using property graphs. Its set of operations can be regarded as a subset of the operations supported by KGTK. To the best of our knowledge, there is no existing toolkit with a comprehensive set of operations for validating, manipulating, merging, and analyzing knowledge graphs comparable to KGTK. 

\section{Conclusions and Future Work}\label{sec:conclusions}

Performing common graph operations on large KGs is challenging for data scientists and knowledge engineers. Recognizing this gap, in this paper we presented the Knowledge Graph ToolKit (KGTK): a data science-centric toolkit to represent, create, transform, enhance, and analyze KGs. KGTK represents graphs in tabular format, and leverages popular libraries developed for data science applications, enabling a wide audience of researchers and developers to easily construct KG pipelines for their applications. KGTK currently supports thirteen common operations, including import/export, filter, join, merge, computation of centrality, and generation of text embeddings. We are using KGTK in our own work for three real-world scenarios which benefit from integration and manipulation of large KGs, such as Wikidata and ConceptNet.

\begin{figure}[t!]
    \centering
    \includegraphics[width=0.9\textwidth]{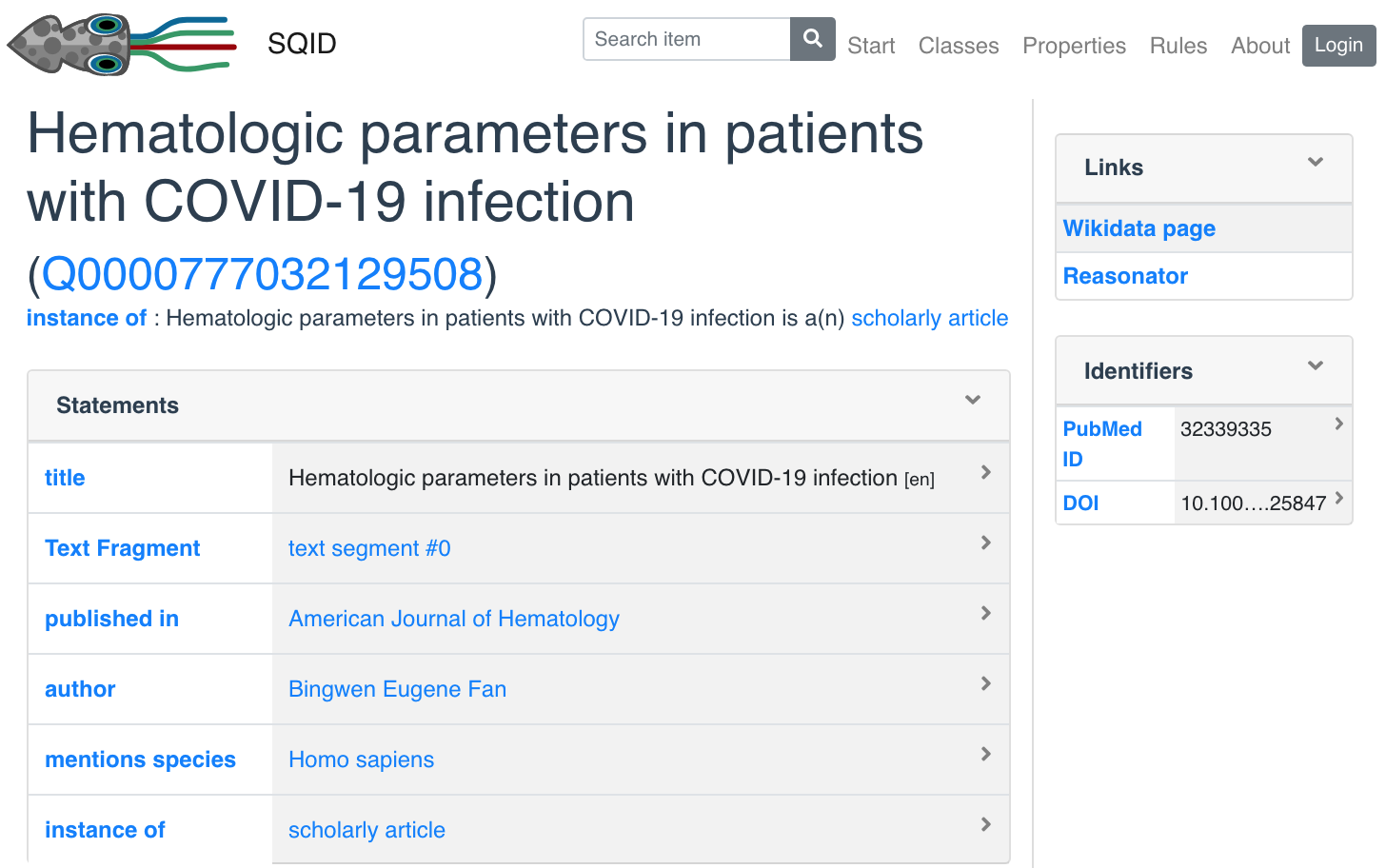}
    \caption{SQID visualization of local KGTK data (using the CORD-19 example).}
    \label{fig:squid}
\end{figure}

KGTK is actively under development, and we are expanding it with new operations. Our CORD-19 use case indicated the need for a tool to create new edges, which will also be beneficial in other domains with emerging information and many  long-tail/emerging new entities. Our commonsense KG use case, which combines a number of initially disconnected graphs, requires new operations that will perform de-duplication of edges in flexible ways. Additional import options are needed to support knowledge sources in custom formats, while new export formats will allow us to leverage a wider span of libraries, e.g., the GraphViz format enables using existing visualization tooling. 
We are also looking at converting other existing KGs to the KGTK format, both to enhance existing KGTK KGs, and to identify the need for additional functionality. In the longer term, we plan to extend the toolkit to support more complex KG operations, such as entity resolution, link prediction, and entity linking.

We are also working on enhancing further the user experience with KGTK. 
We are adapting the SQID\footnote{\url{https://tools.wmflabs.org/sqid/}} KG browser (as shown in Figure \ref{fig:squid}), which is part of the Wikidata tool ecosystem. To this end, we are using the KGTK export operations to convert any KGTK KG to Wikidata format (JSON and RDF as required by SQID), and are modifying SQID to remove its dependencies on Wikidata. The current prototype can browse arbitrary KGTK files. Remaining work includes computing the KG statistics that SQID requires, and 
automating deployment of the Wikidata infrastructure for use with KGTK KGs.

\section*{Acknowledgements}
This material is based on research sponsored by Air Force Research Laboratory under agreement number FA8750-20-2-10002. The U.S. Government is authorized to reproduce and distribute reprints for Governmental purposes notwithstanding any copyright notation thereon. The views and conclusions contained herein are those of the authors and should not be interpreted as necessarily representing the official policies or endorsements, either expressed or implied, of Air Force Research Laboratory or the U.S. Government.





\bibliographystyle{splncs04}

\bibliography{bib}
\end{document}